\pdfoutput=1

\documentclass[11pt]{article}

\usepackage{authblk} 

\usepackage{ACL2023}

\usepackage{times}
\usepackage{latexsym}
\usepackage{graphicx}
\usepackage[T1]{fontenc}

\usepackage[utf8]{inputenc}

\usepackage{microtype}

\usepackage{inconsolata}

\usepackage{booktabs} 

\title{Augmenters at SemEval-2023 Task 1: Enhancing CLIP in Handling Compositionality and Ambiguity for Zero-Shot Visual WSD through Prompt Augmentation and Text-To-Image Diffusion}

\begin{document}

\author[1]{Jie S. Li}
\author[1]{Yow-Ting Shiue}
\author[2]{Yong-Siang Shih}
\author[1]{Jonas Geiping}

\affil[1]{University of Maryland, College Park}
\affil[2]{Duolingo, Inc.}
\affil[ ]{\texttt{\{jli2718,ytshiue,jgeiping\}@cs.umd.edu}}
\affil[ ]{\texttt{yongsiang@duolingo.com}}

\maketitle
\begin{abstract}
This paper describes our zero-shot approaches for the Visual Word Sense Disambiguation (VWSD) Task in English. Our preliminary study shows that the simple approach of matching candidate images with the phrase using CLIP suffers from the many-to-many nature of image-text pairs.
We find that the CLIP text encoder may have limited abilities in capturing the compositionality in natural language. Conversely, the descriptive focus of the phrase varies from instance to instance. We address these issues in our two systems, Augment-CLIP and Stable Diffusion Sampling (SD Sampling). Augment-CLIP augments the text prompt by generating sentences that contain the context phrase with the help of large language models (LLMs). 
We further explore CLIP models in other languages, as the an ambiguous word may be translated into an unambiguous one in the other language. SD Sampling uses text-to-image Stable Diffusion to generate multiple images from the given phrase, increasing the likelihood that a subset of images match the one that paired with the text.

\end{abstract}

\section{Introduction}

The Task of Visual Word Sense Disambiguation, as set out in SemEval-2023 Task 1 Overview Paper \cite{raganato-etal-2023-semeval}, can be described as follows:  Given a target word (the target word) in the context of a two or more word phrase (the full phrase) and ten candidate images, pick the image (the gold image) among the ten candidate images that correctly corresponds to the target word. This competition was run in three languages, English, Farsi and Italian. We participated in the English version of the task. This task is in line with previous tasks connecting images to text, such as \citet{elliott-etal-2017-findings}. We explore two distinct systems to tackle this task. Both systems use Contrastive Language-Image Pre-training (CLIP) \citep{orig_CLIP} as a foundation. CLIP was trained to associate text and related images, through increasing the cosine similarity (CLIP-similarity) between the normalized text-embedding and image-embedding of related text and image pairs and decreasing that for unrelated pairs. Our first system (Augment-CLIP) augments the CLIP text embedding by introducing additional context (through key-to-text) and accessing CLIP text and image embedding in other languages, through third-party implementation of CLIP for various languages. The second system (SD Sampling) samples Stable Diffusion \citep{Stable_Diffusion} to generate multiple images illustrating the semantics of the full phrase and then applies a distance metric to select the candidate image that is closest to the generated images.

As standalone systems, Augment-CLIP and SD Sampling do not outperform Base-CLIP as the additional context may not correctly extend the target word meaning, but they offer complementary benefits and improve Base-CLIP through ensembling. We ensemble models by first calculating a new probability (or score) for each candidate image by taking the equally weighted average of the probability calculated from the underlying models. Each individual model can output a probability for a candidate image. For CLIP-based models, the probability is the softmax of the candidate image logits. We then rank the candidate images based on the new probability in descending order, with the highest probability candidate image being the predicted image from the ensembled model. See Table \ref{tab:Augment-CLIP_performance}.

\section{Systems Overview}

\subsection{Augment-CLIP}
We look at two methods to create the Augment-CLIP system. Both methods attempt to disambiguate the full phrase containing the target word. The first does this through introducing additional text and the second does this through accessing additional languages. 

\subsubsection{Augment-CLIP with key-to-text}
Our baseline approach, referred to as the Base-CLIP approach (or Base-CLIP model) is the approach of encoding the full phrase using the CLIP text encoder and encoding the candidate images using the CLIP image encoder, followed by choosing the candidate image whose encoding has the largest similarity to the full phrase encoding. Base-CLIP models, regardless of their specific underlying architecture, suffer from a weakness in compositionality. Compositionality is the change of word meaning in the presence of other words. For example, the meaning of "baby powder" is not the average of "baby" and "powder", and "powder" means different things in "baby powder" versus "milk powder". This is a general problem with embeddings beyond CLIP, such as text embeddings \citep{embeddings}. CLIP is trained with image and caption pairs only, with captions consisting of shorter texts and of less diversity than texts used to train language models, so the complex syntactic and semantic relationships among words, including compositionality, is not well-captured by CLIP. In comparison, standard language models trained on larger text corpora are composed of longer texts, from a larger variety of sources. We utilize this idea to augment Base-CLIP with key-to-text completion \footnote{\url{https://github.com/gagan3012/keytotext}} to leverage additional language knowledge through the key-to-text system. We use the key-to-text systems "k2t" (k2t 1), "k2t-base" (k2t 2), and "mrm8488/t5-base-finetuned-common\_gen" (k2t 3). 

For example, for the target word "administration" and the full phrase "administration prime minister" from the trial data, we created three additional sets of context texts: 

\begin{enumerate}
    \item "The administration prime minister is the official title of the leader."
    \item "The Administration Prime Minister is the leader of the country."
    \item "prime minister speaks to the media during his visit."
\end{enumerate}

\noindent These texts further reinforce the semantic meaning of "administration". The CLIP text-embedding of the augmented context text is used to measure the CLIP-similarity to the candidate images. 

To keep the focus on the benefit of additional text context rather than optimizing the context itself, we use a greedy method to sample key-to-text and do not evaluate alternative sampling methods. 

\subsubsection{Augment-CLIP through additional languages}
The second way to augment Base-CLIP is to resolve the ambiguity of the full phrase in the source language by translating the full phrase into a different language via a translation model (we leverage Google Translate\footnote{\url{https://translate.google.com/}}) and then use the other language's CLIP text-embedding of the translation to measure the distance to the candidate images. We evaluate this idea with Chinese translations. Chinese Augment-CLIP does not outperform Base-CLIP, often due to poor translation, but, interestingly, it offers sufficient complementarity to Base-CLIP or other Augment-CLIP that it improves performance through ensembling. See results in Table \ref{tab:Augment-CLIP_performance}.   

\begin{table*}[h]
\centering
\footnotesize
\caption{Augment-CLIP systems performance.}
\label{tab:Augment-CLIP_performance}
\begin{tabular}{c|cccc}
\hline
 & Trial Data & & Test Data & \\
Systems &  hit rate &mrr & hit rate & mrr\\
\hline
Base-CLIP (ViT-B/32) & 56.25 & 73.44 & 57.88 & 72.43\\
Aug-CLIP: zh & 43.75 & 62.14 & 51.40  & 67.60\\
Aug-CLIP: k2t 1     & 50.00 & 68.75 & 57.24 & 71.71\\
Aug-CLIP: k2t 2     & 56.25 & 70.42 &58.10 & 71.98\\
Aug-CLIP: k2t 3     & 43.75 & 68.08 & 53.78 & 69.07\\
submit1=ensemble(B-CLIP, k2t 2) & 62.50 & 76.56 & 59.18 & 73.21 \\
ensemble(B-CLIP, zh, k2t 2) & \textbf{68.75} & \textbf{81.25} & 63.71 & 76.11\\
ensemble(B-CLIP, zh, k2t 1, k2t 2, k2t 3) & \textbf{68.75} & \textbf{81.25} &\textbf{63.93}    & \textbf{76.45} \\
\hline
baseline: & & \\
Base-CLIP (ViT-Large-Patch14-336) & 56.25 & 73.12 & 60.48 & 73.88\\
\hline
\end{tabular}
\end{table*}

\subsubsection{Base-CLIP model differences}
For Base-CLIP, the performance difference in the two versions of CLIP that we used, ViT-B/32 and Vit-L/14, is notable. ViT-B/32 in fact gave better performance on trial and test data.  This is unexpected as ViT-L/14 is a larger model and has more training and more data  \citep{orig_CLIP}. Further the organizers' baseline uses CLIP-ViT-large-patch14-336, an even larger model which improved performance in test data. See Table \ref{tab:Augment-CLIP_performance}. This leads to the question of how different Base-CLIP embeddings affect performance on this task, which is outside the scope of this paper as we take the Base-CLIP embedding as a given in our systems. 
 
\subsection{Stable Diffusion Sampling}

The second system samples text-to-image Stable Diffusion-v1.4 (SD) to generate multiple images after inputting the full phrase as the text prompt. Then the system outputs the candidate image with the closest distance to any of the generated SD images. There are two advantages of this system: first is the access to the larger training data of Stable Diffusion, which includes LAION2B-en \citep{Schuhmann2022LAION5BAO}, a 2.32 billion common crawl image-text pairs dataset. Second, evaluating multiple images for a given text input resolves the text ambiguity of the input text and also the pictorial ambiguity in its image representation. As an example of text ambiguity, "angora" can mean a type of fiber or less frequently a specific city, as in "Angora City". Sampling several images allows the possibility that a subset of the images correctly express the meaning of the target word. Even for an unambiguous word, there may be pictorial diversity in its representation, and sampling multiple images allows for broader coverage of this diversity than a single image. 

We evaluate two sampling methods of Stable Diffusion, text-to-image and text-and-image-to-image. For each, two similarity metrics were used: CLIP-similarity and $l_2$ distance between InceptionV3 \citep{InceptionV3} features of candidate image and InceptionV3 features of SD sampled image. Of these four, text-to-image sampling of Stable Diffusion with CLIP-similarity performs the best on trial data and a subset of train data - this is designated SD Sampling and is our submission 2 for the task. 

For text-to-image sampling, we input the full phrase to SD and generate 50 output images (independent of any candidate images). We then calculate the maximum CLIP-similarity (CLIP ViT-L/14) between a candidate image and the 50 output images and associate that largest CLIP-similarity to that candidate image (candidate image distance). We then output the candidate image with the largest CLIP-similarity.

\section{Experimental setup}

The trial, train, and test datasets consist of multiple instances. An instance is a target word and a full phrase (containing the target word) and ten candidate images, with one image (the gold image) capturing the semantic meaning of the target word as exemplified in the full phrase. Train, trial, and test have $12869$, $16$, and $463$ instances, respectively. For the test data, there are two versions of the dataset provided by the task organizers, differing in the image size \cite{raganato-etal-2023-semeval}. We perform our predictions on the dataset with the smaller image size.

We do not train or fine-tune our models on training data to demonstrate the zero-shot property of our approach, although we do use the training data in part to inform us of which Augment-CLIP system and which SD Sampling system to select for task submissions. Based on trial data performance, among the three k2t systems, we choose k2t 2, and among the SD Sampling systems, we choose text-to-image with CLIP-similarity.

As measurements of the performance of the models, hit rate and mean reciprocal rank (mrr) are applied to the model predictions on the trial dataset and test dataset. Based on the inputs, the model assigns a score to each candidate image. The model can output one predicted image, with the highest score, or it can output a list of images ordered in decreasing order of score. Hit rate is the percentage of instances where the predicted image is the gold image. Mean reciprocal rank is the average of the reciprocal of the rank of the gold image in the list of images, ordered based on score. 

\section{Results}

\subsection{Augment-CLIP through key-to-text}
While standalone Augment-CLIP through key-to-text does not outperform Base-CLIP, it does reveal that adding context can improve performance. The additional context, when correctly augmenting the meaning of the target word, can indeed improve performance on the test set. In the best-case scenario, the additional context is an extension and explanation of meaning of the target word. In the worst-case, an incorrect extension of context dilutes the meaning of the target word. In the former case, Augment-CLIP is likely to correctly predict the gold image. In the group of instances in which both Augment-CLIP and Base-CLIP correctly predict the gold image,  the CLIP-similarity score is higher in Augment-CLIP than in Base-CLIP, showing the effectiveness of added context. This depends on the quality of the context extension process: if the augmented context does not aid in conveying the correct semantic meaning of the target word, then the incorrect additional context may degrade performance in a standalone system. This is analogous to the performance of a language model with in-context learning, where the performance depend on the quality of in-context examples \citep{inContextLearning}.

\begin{table}[t]
    \centering
    \caption{Mean of CLIP similarity values of an instance of the full phrase to gold image versus text to all candidate images in the test data. Given the scale of similarity values shown in this table, increasing the similarity value between text and its associated image by values as small as $0.01$ could noticeably improve the system performance. }
    \footnotesize
    \label{tab:base_augment_sim_mean}
    \vspace{0.1in}
    \begin{tabular}{c|cc}
    \hline
     & Test Data & Test Data \\
        Systems     & sim(text, gold img) & sim(text, all imgs) \\
    \hline
        Base-CLIP           & 0.2932 & 0.2276 \\
        Aug-CLIP: kt2 2 & 0.2929 & 0.2276 \\
    \hline
    \end{tabular}
\end{table}

Adding a k2t system can improve performance of the Base-CLIP. This can be seen in Table~\ref{tab:k2t_sim_difference}. For each instance in the dataset, consider the Base-CLIP similarity score for the full phrase and the gold image and consider the Augment-CLIP through k2t similarity score for the full phrase and the gold image. The difference between the Augment-CLIP similarity score and the Base-CLIP similarity score is calculated and shown in Table~\ref{tab:k2t_sim_difference}. This difference shows whether Augment-CLIP would have done better or worse than Base-CLIP. It also shows the potential of Augment-CLIP to improve Base-CLIP's performance.

Extra steps can be taken to improve the quality of the k2t text completion, but our focus is not to improve the performance of the k2t system but to show that reasonable additional context offers complementary benefits to Base-CLIP. 

\begin{table}[h]
\centering
\footnotesize
\caption{Test dataset confusion matrix: Aug-CLIP k2t 2 similarity(text, gold image) minus Base-CLIP similarity(text, gold image) and count of instances. Aug-CLIP can increase performance of Base-CLIP when Base-CLIP is incorrect by increasing the similarity(text, gold image).}
\label{tab:k2t_sim_difference}
\begin{tabular}{cc|cc}
\toprule
& difference & Aug-CLIP: k2t 2 &\\
& in similarity & correct &  incorrect\\
\midrule
Base-CLIP & correct           & 0.0015 & -0.0248\\
&incorrect             & 0.0107 & -0.0025 \\ 
\hline
\hline
 & instance & Aug-CLIP: k2t 2 &\\
& count & correct &  incorrect\\
\midrule
Base-CLIP & correct           & 234 & 34 \\
& incorrect             & 36 & 159 \\ 
\hline

\end{tabular}
\end{table}

\subsection{Augment-CLIP through other languages}

We evaluate another method to disambiguate the full phrase by translating the full phrase in English to another language and exploring the CLIP text embedding and image embedding in that foreign language. Direct translation to a foreign language (through taking the first result of Google Translate), with that language chosen to be Chinese (Aug-CLIP: zh) in our evaluation, does not increase performance, and this is partly due to incorrect translations. Here, identical round-trip translations can serve as a proxy for correct translation from English to Chinese. The test instances can be divided into two groups, one being identical in round-trip translation and the other group containing all other instances. For instance, starting with the English full phrase and translating to Chinese and then translating that result back to English (English\_1 $\rightarrow$ Chinese  $\rightarrow$  English\_2 with English\_1 and English\_2 being the same, up to capitalization) and another group that does not have identical round-trip translation, the first group has a higher foreign-language CLIP-similarity score than the second. 

As a standalone system, direct translation does not improve performance, but ensembling with a direct translation system does improve performance. By adding Chinese translation to the ensemble (ensemble(B-CLIP, zh, k2t 2)), test data hit rate increases from $59.18$ to $63.71$ and test data mrr increases from $73.21$ to $76.11$. See Table \ref{tab:Augment-CLIP_performance}. 

\subsection{SD Sampling}

SD Sampling does not outperform Base-CLIP. It's worth noting that the instances where SD Sampling correctly selects the gold image is different from those of Base-CLIP, showing a potential to gain from accessing the SD Sampling system. See Table \ref{tab:SDvBaseClipConfusionMatrix}.

\begin{table}[h]
\centering
\footnotesize
\caption{SD Sampling versus Base-CLIP test dataset confusion matrix: count of test instances.}
\label{tab:SDvBaseClipConfusionMatrix}
\begin{tabular}{cc|cc}
\toprule
& instance & SD Sampling &\\
& count & correct &  incorrect\\
\midrule
Base-CLIP & correct           & 190 & 78 \\
& incorrect             & 56 & 139 \\ 
\hline

\end{tabular}
\end{table}

\begin{figure}
    \centering
    \includegraphics[width=8cm]{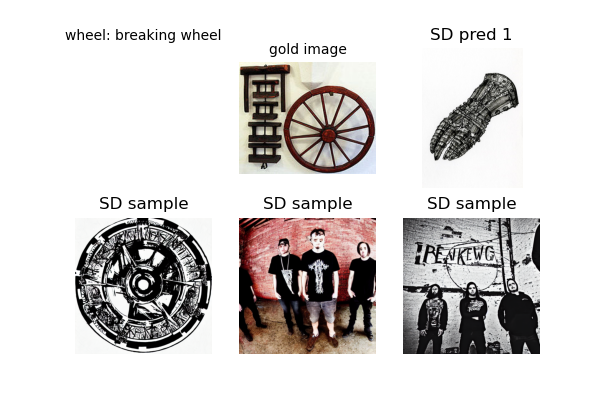} 
    \caption{Top row: "wheel" in "breaking wheel" with gold image and SD Sampling predicted candidate image. Bottom row: Examples from SD Sampling of "breaking wheel", as both an object and a music band.} 
    \label{fig:bwheel}
\end{figure}

\begin{figure}
    \centering
    \includegraphics[width=8cm]{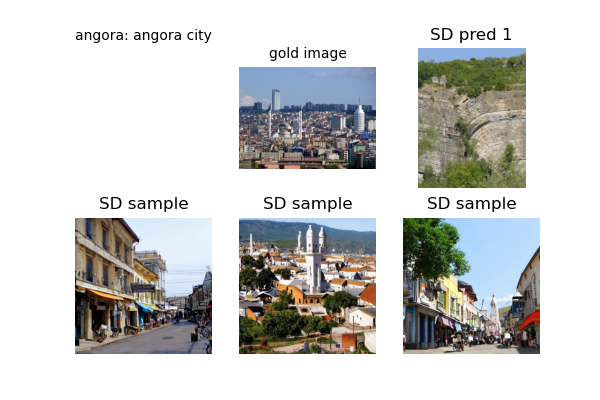} 
    \caption{Top row: "angora" in "angora city" with gold image and SD Sampling predicted candidate image. Bottom row: Examples from SD Sampling, with various views of a city. A view of a side street of a city matches more closely an incorrect candidate image of a natural wall.} 
    \label{fig:angoraCity}
\end{figure}

\begin{figure}
    \centering
    \includegraphics[width=8cm]{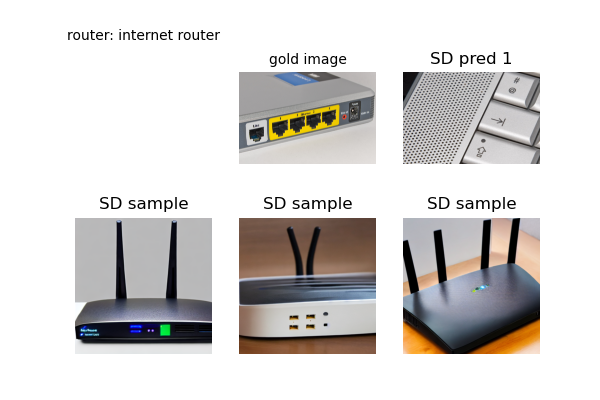} 
    \caption{Top row: "router" in "internet router" with gold image and SD Sampling predicted candidate image. Bottom row: Examples from SD Sampling, with various views of a router. A broad view of a router does not match the close up view presented in the gold image.} 
    \label{fig:internetRouter}
\end{figure}

There is pictorial diversity in the SD samples, and often that diversity includes the correct image expression of the target word in the full phrase, as intended. There is diversity in viewpoint, proximity and style of the object presented. See images of various cityscapes outputted by Stable Diffusion for the full phrase "angora city" in Figure \ref{fig:angoraCity}, and see images of various views of different models of "internet routers" in Figure \ref{fig:internetRouter}. There is also diversity in the semantic interpretation of the full phrase: see for example Figure \ref{fig:bwheel} for interpretation of the word "breaking wheel" as both a torture device and a music group. This shows that the first goal of the SD System of producing a diversity of pictorial representation of the desired object is met. But the subsequent application of the distance metric fails to match the sampled SD image to the gold image. At times, incorrect candidate images have larger CLIP-similarity to the correct sampled SD image than the gold image, due to a coincidence of similar style, lighting, or material. This is not a shortcoming of CLIP-similarity as it is intended to be applied to (text, image) pairs and not (image, image) pairs \citep{orig_CLIP}. As an alternative, we evaluate metrics such as $l_2$ distance between InceptionV3 features of the sampled image and InceptionV3 features of the candidate image. Using $l_2$ metric underperforms the CLIP-similarity metric as shown Table \ref{tab:SD_trial_train_performance}. We do not evaluate other image-to-image similarity metrics and leave for future work the search for an effective metric.

An issue with SD sampling on this dataset is the domain shift between the dataset on which SD was trained, a common crawl of text and image pairs in English, and the more scientific and technical focus of the full phrases in the test data. For example, the full phrase "breaking wheel" is a historical term and meant to be unambiguous and to resolve to mean a medieval torture device and the gold image is of such a device. On the other hand, to the layperson, "breaking wheel" sounds like the name of a band, akin to Stone Sour, Breaking Benjamin, Nickelback, and this popular understanding of "breaking wheel" is evidenced in Stable Diffusion sampled images, which include images of band groups. Similarly, other instances with the full phrase being technical and scientific terms that are not well-known to the general public are expressed in Stable Diffusion output images in terms of how a layperson would interpret such a term, instead of the correct technical meaning.  

\begin{table}
\centering
\footnotesize
\caption{SD Sampling systems performance.}
\label{tab:SD_trial_train_performance}
\begin{tabular}{c|cc}
\hline
 & Trial Data & \\
Systems &  hit rate &mrr \\
\hline
(text)2img inception    & 31.25 & 48.24    \\
(text,img)2img inception & 31.25 & 53.39   \\
submit2=(text)2img CLIP & 31.25 & \textbf{56.00}    \\
(text,img)2img CLIP     & \textbf{37.50} &  55.00   \\
\hline
\hline
 & Test Data & \\
Systems &   hit rate & mrr\\
\hline
(text)2img inception        & 34.77 & 52.59 \\
(text,img)2img inception   & 28.08 & 49.41 \\
submit2=(text)2img CLIP    & \textbf{53.35} & \textbf{68.75} \\
(text,img)2img CLIP       & 44.28 & 63.14 \\
\hline

\hline
\end{tabular}
\end{table}

\section{Conclusion}
The Base-CLIP system is a strong solution to the task challenge. 
Our system Augment-CLIP complements Base-CLIP to resolve text ambiguity and improves text compositionality. Our system SD Sampling provides pictorial diversity in ambiguous and unambiguous text interpretation. These two methods offer additional ways to connect text and images.  

\newpage

\bibliography{custom}
\bibliographystyle{acl_natbib}

\appendix

\label{sec:appendix}

\end{document}